\documentclass[preprints,article,accept,moreauthors,pdftex]{Definitions/mdpi}

\firstpage{1} 
\pubvolume{1}
\issuenum{1}
\articlenumber{0}
\pubyear{2024}
\copyrightyear{2024}
\datereceived{ } 
\daterevised{ } 
\dateaccepted{ } 
\datepublished{ } 
\hreflink{https://doi.org/} 
\pdfoutput=1 

\Title{Reservoir Computing Using Measurement-Controlled Quantum Dynamics}

\TitleCitation{Reservoir Computing Using Measurement-Controlled Quantum Dynamics}

\definecolor{my_green}{rgb}{0.55, 0.71, 0.0}

\usepackage{textcomp}
\usepackage{bm}
\usepackage{comment}

\Author{A.~H.~Abbas $^{*}$\orcidA{} and Ivan S.~Maksymov\orcidB{}}

\AuthorNames{Firstname Lastname, Firstname Lastname and Firstname Lastname}

\AuthorCitation{A.~H.~Abbas; Ivan S.~Maksymov.}

\address{%
Artificial Intelligence and Cyber Futures Institute, Charles Sturt University, Bathurst, NSW 2795, Australia.}

\corres{Corresponding author:~aborae@csu.edu.au}

\abstract{Physical reservoir computing (RC) is a machine learning algorithm that employs the dynamics of a physical system to forecast highly nonlinear and chaotic phenomena. In this paper, we introduce a quantum RC system that employs the dynamics of a probed atom in a cavity. The atom experiences coherent driving at a particular rate, leading to a measurement-controlled quantum evolution. The proposed quantum reservoir can make fast and reliable forecasts using a small number of artificial neurons compared with the traditional RC algorithm. We theoretically validate the operation of the reservoir, demonstrating its potential to be used in error-tolerant applications, where approximate computing approaches may be used to make feasible forecasts in conditions of limited computational and energy resources.}


\begin{document}


\section{Introduction}
As the semiconductor electronics approaches its fundamental and technological limitations, the research community is increasingly turning its attention to novel computation paradigms to enable further improvement of computers. Modern digital computers can solve virtually any computational problem. However, to accomplish a computational task of arbitrary complexity, they may require impracticably large resources such as time and memory. To resolve this challenge, unconventional \cite{Ada17, Ada19} and neuromorphic \cite{Sha20, Mar20, Mar20_1, Sua21, Rao22, Sar22, Sch22, Kra23} computing were proposed as the new methods of computer engineering, where elements of a computer mimic the operation of a biological brain relying on physical and chemical processes \cite{Nak21, Mak23_review}.

While neuromorphic computers may not be as universal as the traditional digital ones, they can solve certain practically important problems with feasible accuracy using just a small amount of computational resources and energy needed by a high-performance computer tasked with the same problem. Neuromorphic computers are also inherently scalable, parallel and allow for collocation of data processing and memory \cite{Sch22}. Similarly to a biological brain, they also operate only when input data are available and mimic the randomness of the firing of biological neurons, thus helping save energy and decrease the overall cost of computations \cite{Maa02, Jae04}. 

These qualities make the neuromorphic computers ideally suitable for applications in approximate computing, another emergent approach to computations that benefits the fields of machine learning, multimedia processing, signal processing and scientific computing by replacing high-accuracy resource- and energy-consuming computations by alternative solutions that produce practicable results using less energy and resources \cite{Mit16, Liu20, Hen22, Ull23, Mak21_ESN}. Importantly, both approximate computing and neuromorphic computing approaches also utilise errors as an opportunity for enhancing efficiency, mimicking the ability of a biological brain to learn and improve from errors \cite{Sch22}.

Reservoir computing (RC) is a resource-efficient neuromorphic computing algorithm that is especially suitable for making forecasts of highly nonlinear and chaotic time series that underpin a number of essential natural and human-made phenomena, including the variation of climate, dynamics of Earth population, trends in financial markets, energy generation and drug discovery \cite{Luk09, Luk12, Bal18, Tan19, Nak20, Cuc22, Dam22, Zha23, Mak23_review}. A typical RC algorithm \cite{Luk09, Luk12} employs a randomly initialised task-dependent neural network (reservoir) that is connected to the input units through random connections. The dynamics of the reservoir is advanced in time using a nonlinear update equation, resulting in a set of activation states. Then, output weights are computed as the linear regression weights of the teacher outputs on the activation states. The so-trained RC system is then tasked to either solve a classification problem or make forecasts using a new set of input data.

A computational reservoir can also be created using a physical, either experimental or theoretical, nonlinear dynamical system that effectively recreates the dynamical properties of the update equation of the traditional RC algorithm \cite{Luk09}. Called the physical reservoir \cite{Nak21}, this approach to computer engineering has been successfully validated using spintronic devices \cite{Rio19, Wat20, All23}, electronic circuits \cite{Cao22}, photonic systems \cite{Sor20, Raf20}, mechanical devices \cite{Cou17} and liquids \cite{Khe22, Gao22, Mak23_review, Mar23}. 

Yet, similarly to the advantage of quantum computers over classical digital computers \cite{Nie02}, quantum physical RC systems---neuromorphic computers with a reservoir operating according to the laws of quantum mechanics \cite{Nak21, Muj21, Gov21, Suz22, Gov22_1, Dud23, Got23, Llo23, Cin24}---offer a number of advantages over classical, both algorithmic and physical, RC systems. In particular, quantum physical RC systems have demonstrated a superior capability of predicting complex dynamical systems with many degrees of freedom \cite{Got23, Cin24} and the ability to create a large number of densely connected artificial neurons using technically simple coupled quantum oscillators instead of sophisticated physically coupled qubits \cite{Dud23}. Moreover, it has been demonstrated that while the quantum noise is undesirable in conventional quantum computations \cite{Nie02} it can be exploited as a computational reservoir \cite{Suz22}.

In this paper, we propose and theoretically validate a novel quantum RC architecture that exploits the dynamics of an atom trapped in a cavity. One prominent feature of the proposed system is a coherent driving of the atom at a certain driving rate with the possibility to observe transitions between quantum states and effective ``freezing'' of the quantum evolution of the system. Frequent observation of the atom eigenstate leads to the prevention of the system from undergoing significant changes, a phenomenon known as the Zeno effect \cite{Harr17}. Conversely, less frequent probing of the system states enables the system to undergo Rabi oscillations \cite{Rai12, Lew23}. Judiciously using these properties, we optimise the rate at which the atom is driven (i.e.~we optimise the measurement rate of atomic eigenstates) to adjust the quantum dynamics of the reservoir to undertake diverse classification and prediction tasks. This approach opens up opportunities for controlling and stabilising quantum states during a computation, benefiting such essential applications as decoherence mitigation \cite{Yas16}, quantum information processing \cite{Ale09}, quantum error correction and quantum state stabilization \cite{Paz12, Lew23, Bur23}.

The remaining discussion is structured as follows. In Section~\ref{sec:Theory}, we theoretically describe the dynamics of a probed atom in a cavity subjected to coherent detection. In Section~\ref{sec:QRM}, we introduce the foundations of the quantum reservoir model. Then, in Section~\ref{sec:Test} we test the reservoir on a number of challenging benchmarking tasks, including a classification task, chaotic time series free-running forecast task and physical system prediction task. We also highlight the role of the model parameters on the accuracy of forecasts made by the reservoir, including the effect of the cavity driving amplitude, number of neurons, the measurement rate of the atomic state and length of the training datasets. Finally, in line with the envisioned applications of the reservoir model in the field of approximate computing, we demonstrate that the use of the measurement-controlled dynamics significantly decreases the amount of computational resources required by the previously proposed quantum RC systems to successfully undertake complex prediction tasks. 
\section{Atom-cavity interaction under measurement control\label{sec:Theory}}
We consider a physical system that consists of a two-level atom (qubit)  that is trapped inside a cavity. The qubit is represented by a mode $\sigma$ but the cavity is represented by a mode $a$. In the framework of this simplification, the model is entirely isolated from any external influences, allowing only for measurement-controlled interactions of the qubit with a coherent field input and accounting for an inherent decay of the cavity.

In our model, the atom has two possible spin states $|0\rangle=|\uparrow\rangle$ and $|1\rangle=|\downarrow\rangle$ that interact with a quantised electromagnetic field within a cavity. The governing Hamiltonian $\hat{H}_i$ for this atom-cavity interaction is expressed as 
\begin{eqnarray}
\hat{H}_i = g a^\dagger a \sigma_{-} \sigma_{+}\,,
\label{eq:RC2}
\end{eqnarray}
where $g$ represents the strength of the atom-cavity coupling, $a$ is the cavity annihilation operator and $\sigma_{-}$ and $\sigma_{+}$ are the lowering and raising operators for the atom, respectively. The interaction between the atom and the cavity leads to exchange of energy, giving rise to Rabi oscillations---oscillations of energy between the atom and the cavity. Yet, the cavity ensures multiple reflection of the probe field, thereby effectively enhancing the strength of the interaction between the electromagnetic field and the atom. The cavity also undergoes coherently driven through one mirror. This process is described by the Hamiltonian 
\begin{eqnarray}
\hat{H}_c = -i\beta(a^\dagger - a)\,,
\label{eq:RC3}
\end{eqnarray}
where $\beta$ represents the cavity driving amplitude, serving as the carrier for input data injected into the reservoir as we will discuss later.

Importantly, the state of the atom is monitored through a coherent continuous measurement. The corresponding Hamiltonian for this process is given by
\begin{eqnarray}
\hat{H}_z = g_z (\sigma_{+} + \sigma_{-})\,,
\label{eq:RC4}
\end{eqnarray}
where $g_z$ represents the amplitude of the coherent atomic drive. Adjusting the parameters of the atom-cavity interaction and incorporating the atomic driving parameter $g_z$ into the model, we can manipulate the evolution of the quantum system.

\begin{figure}[H]
\centering
   \includegraphics[width=0.7\columnwidth]{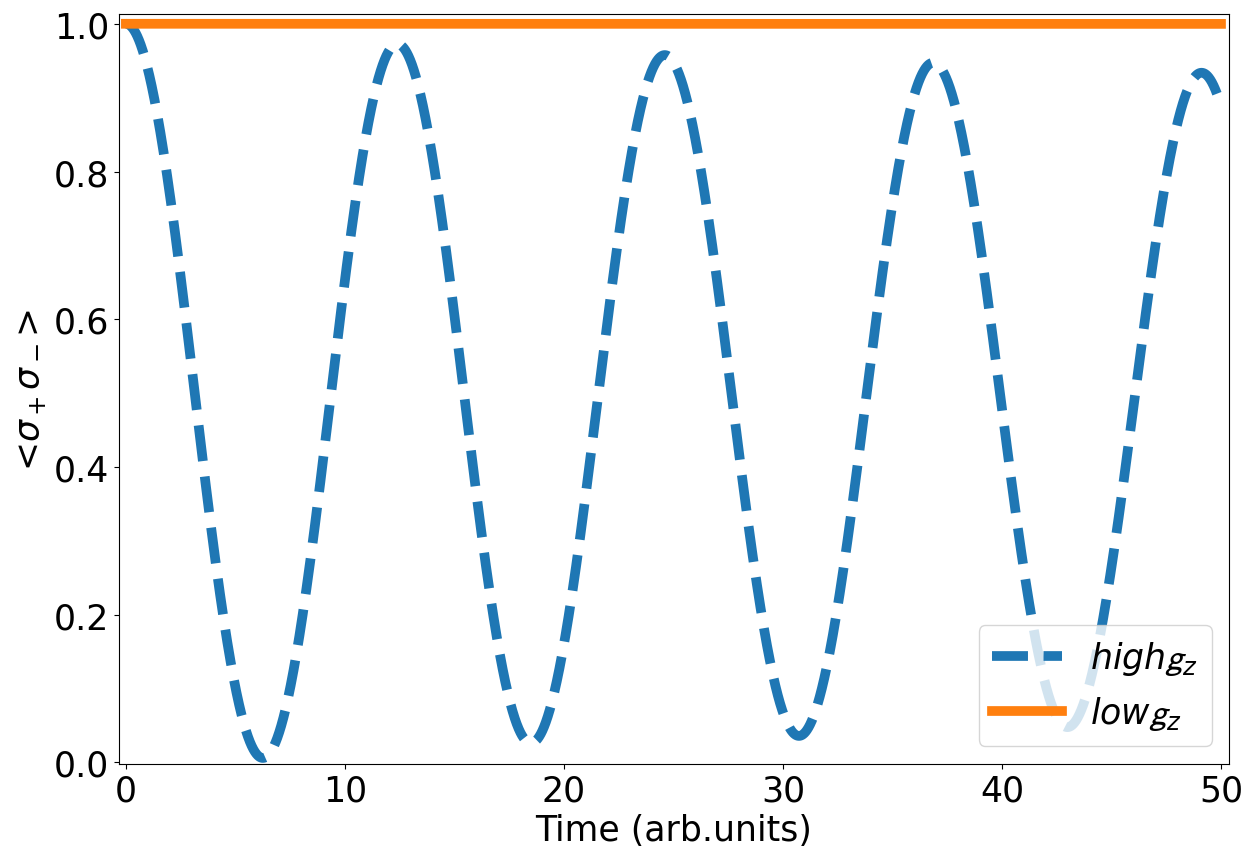}
\caption{Time evolution of the operator $\langle\sigma_+ \sigma_-\rangle$ for a high (infrequent measurement) and low (frequent measurement) values of $g_z$. Depending on the measurement rate, the system exhibits a Zeno effect with frequent measurements (the orange solid line) or it undergoes oscillations when the measurement rate is low (the blue dashed line).}
\label{f1-1}
\end{figure}

Figure~\ref{f1-1} illustrates the time evolution of the operator $\langle\sigma_+ \sigma_-\rangle$. Depending on the driving amplitude of the atom $g_z$, the system either exhibits a freezing behavior akin to the Zeno effect with frequent measurements (depicted by the solid orange line) or it undergoes oscillations (represented by the dashed blue line). In particular, while the atom-cavity interaction provides a natural platform for quantum manipulation, controlling the driving amplitude of the atom enables us to tune the evolution of the system. In our model, the parameter $g_z$ defines the measurement rate of the atomic states. Henceforth, we will employ the term ``measurement rate'' to denote the driving amplitude of the atom, represented by $g_z$.
\subsection{System dynamics}
We analyse the dynamics of the proposed RC system using a stochastic master equation approach where the system is represented as a qubit and a cavity mode \cite{Nie08}. The master equation considered by us corresponds to the time-dependent Schr{\"o}dinger equation that accounts for the atom-light interaction and the effect of frequent measurements on the spin states of the atom. The full Hamiltonian for the system is
\begin{eqnarray}
\hat{H} = \hat{H}_i +\hat{H}_c +\hat{H}_z\,, 
\label{eq:RC6}
\end{eqnarray}
where $\hat{H}_i$, $\hat{H}_c$ and $\hat{H}_z$ are given by Eqs.~\eqref{eq:RC2},~\eqref{eq:RC3}~and~\eqref{eq:RC4}, respectively.
The time evolution of the density matrix $\rho$ is governed by the linear stochastic master equation, which accounts for the effects of decoherence and dissipation.
\begin{eqnarray}
\dot{\rho }=-i[\hat{H},\rho ]+\hat{C}\rho {\hat{C}}^{{\dagger} }-\frac{1}{2}{\hat{C}}^{{\dagger} }\hat{C}\rho -\frac{1}{2}\rho {\hat{C}}^{{\dagger} }\hat{C}\,,
\label{eq:RC7}
\end{eqnarray}
where $\hat{H}$ is given by Eq.~\eqref{eq:RC6} and $\hat{C} = \sqrt{\kappa} a$ is the collapse operator associated with the cavity decay and $\kappa$ is the decay rate. By numerically solving Eq.~\eqref{eq:RC7} we simulate measurement-driven dynamics of a trapped atom in a cavity.

\begin{figure}[H]
\centering
   \includegraphics[width=0.77\columnwidth]{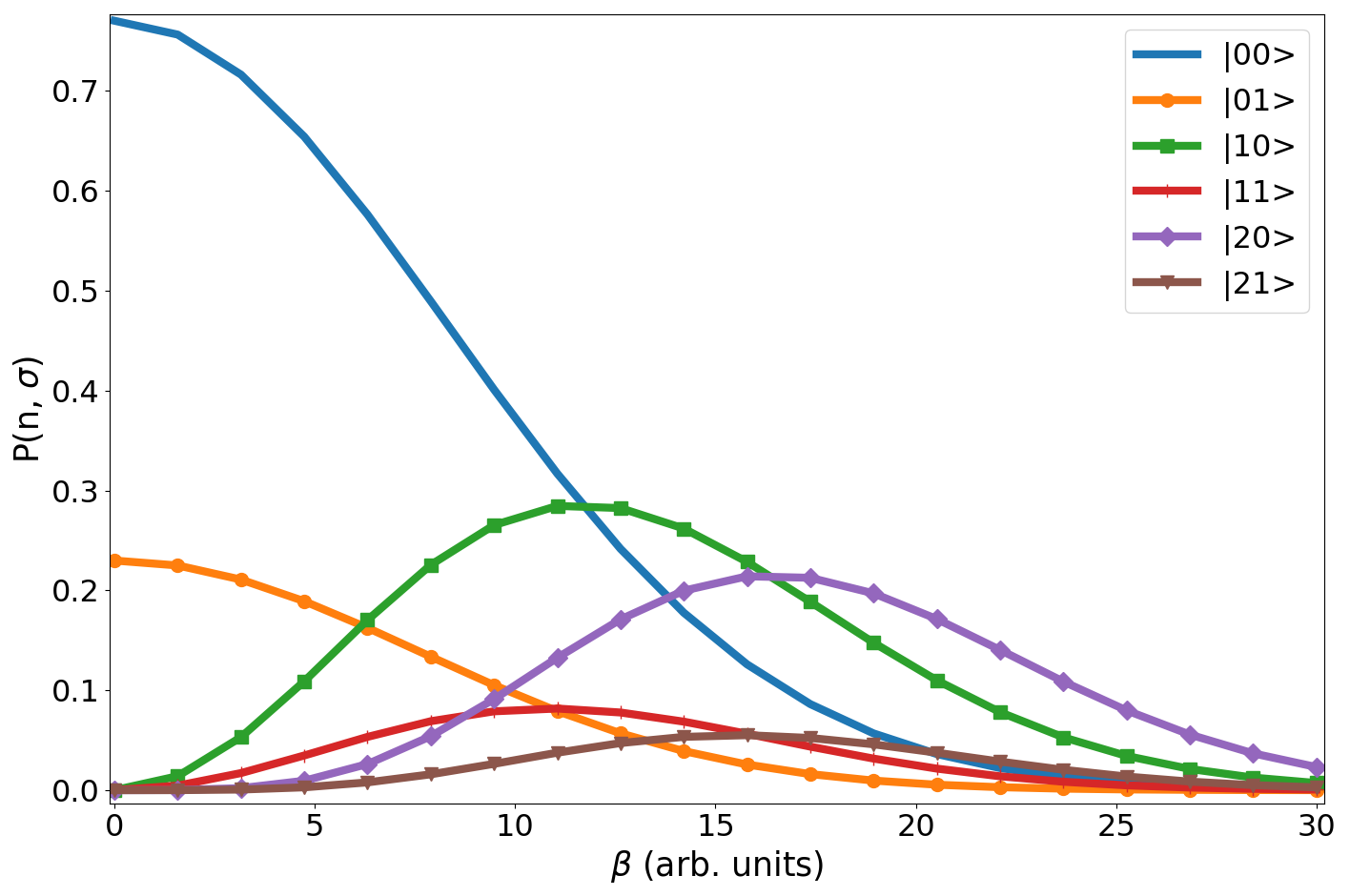}
\caption{ Occupation probabilities of Fock states $|n\sigma\rangle \in |00\rangle\dots|21\rangle$ as a function of cavity driving amplitude $\beta$.}
\label{f2}
\end{figure}

The basis of the quantum states within the cavity $ |n, \sigma \rangle$ is generated by the tensor product of an atom with the two possible spin states and the $n$-dimensional space of a quantised field ($|n, \sigma \rangle = |n \rangle \otimes |\sigma\rangle$). The population of the Fock states is driven by the coherent amplitude ($\beta$) and is further manipulated by means of the measurement rate $g_z$. The occupation probabilities of these states are given by
\begin{eqnarray}
P(n,\sigma) = \langle n\sigma |\rho|n\sigma\rangle\,,
\label{eq:RC7_1}
\end{eqnarray}
where $\rho$ is the density matrix, whose time derivative is given by Eq.~\eqref{eq:RC7}. To optimize the operation of the reservoir, the cavity driving amplitude $\beta$ is adjusted to initially populate the first few Fock states with a significant probability. In Figure~\ref{f2}, the occupation probabilities \(P(n,\sigma)\) for the first six Fock states are depicted as a function of $\beta$. These probabilities exhibit a nonlinear response and, for values of $\beta$ within the range $[10, 15]$, there is a non-zero occupation probability for all Fock bases $|n\sigma\rangle \in |00\rangle\dots|21\rangle$.

\section{Quantum Reservoir Model\label{sec:QRM}}
\begin{figure}[H]
\centering
\includegraphics[width=0.8\columnwidth]{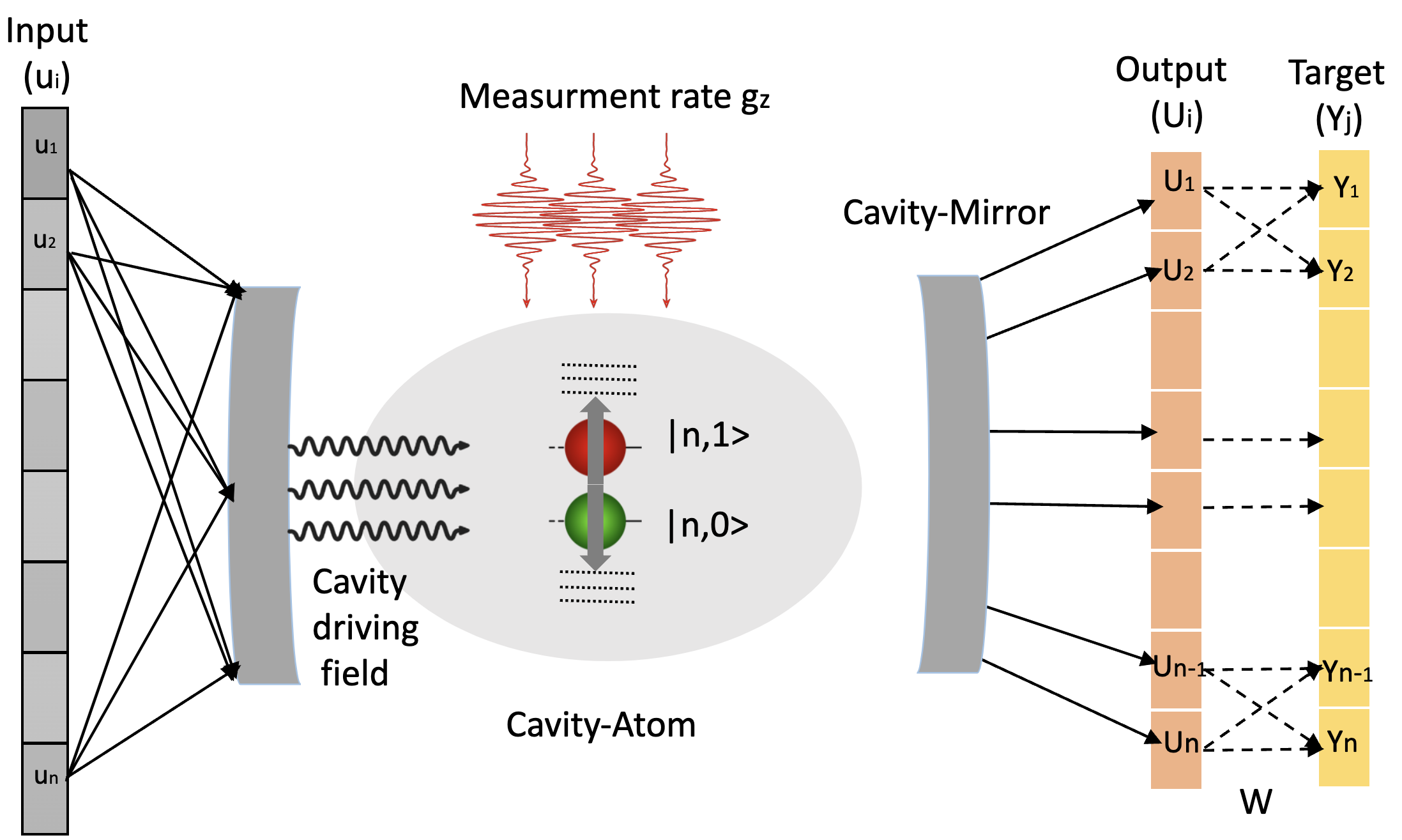}
\caption{Sketch of the RC system with a measurement-controlled quantum dynamics. The pivotal component of the reservoir is a cavity-atom system with continuously monitored states. The neural activations of the reservoir are given by the Fock states of the atom-cavity quantum system. The reservoir is coherently driven using a signal defined by the discrete points of the input dataset. The classified readouts of the reservoir are processed by means of a linear regression technique.} 
\label{f1}
\end{figure}

The principle of operation of the quantum reservoir is illustrated in Figure~\ref{f1}. The neurons of the reservoir are defined by Fock states $|n, \sigma \rangle$ that form the basis for the quantum states within the cavity and constituting a computational space of $2n$ states. The input data points are converted into signals that modulate the driving amplitude $\beta(t)$ used to change the distribution of the Fock states population (effectively, as explained above, the reservoir nonlinearly transforms the input data into a high-dimensional state space). The input information are carried by electromagnetic radiation delivered through one mirror of the cavity and output detection is performed at the opposite cavity side. The outputs of the reservoir are determined by the expectation values $P(n,\sigma)$ of the occupancy of the basis states. These occupation probabilities are subjected to classification through a trainable fully connected layer (illustrated by the dashed black arrows in Figure~\ref{f1}) using a regression method. 

Moreover, we adjust the dynamics of the reservoir to a specific task using the measurement rate. That is, in our system the measurement rate effectively corresponds to the leaking rate parameter that controls the dynamics of the traditional RC algorithm and that needs to be adjusted for every specific problem \cite{Luk09, Luk12}. For instance, when the quantum reservoir is tasked with a problem that involves an extended plateau regime, its dynamics can be slowed down (``frozen'') using a suitable value of $g_z$. Conversely, the dynamics of the reservoir can be accelerated when the input of the RC system is defined by a rapidly evolving time-series dataset.

The reservoir is trained using the input dataset $u = \{u_1 , u_2 , \ldots, u_{_M} \}$ with each data point corresponding to a discrete instant of time $t_{_M}$. Additionally, each point of the input dataset is associated with a training dataset ${\tilde{{\bf{Y}}}}_{\rm{train}}$. As part of an iterative procedure,  the input values $u_i$ with $i=1 \dots M$ are encoded into a the driving amplitude $\beta$ that is used to drive the dynamics of the reservoir (this computation corresponds to the solution of Eq.~\eqref{eq:RC7} with the Hamiltonian given by Eq.~\eqref{eq:RC6}). The values of the output layer ${\bf{U}}$ are obtained by measuring a set of observables $P(n,\sigma)$. The linear combination of these observables is then optimised to generate the target associated with each task.

Thus, the output of the reservoir can be represented as
\begin{eqnarray}
{\tilde{{\bf{Y}}}}= {\bf{W}} {\bf{F}}({\bf{U}})\,,
\label{eq:RC8_1}
\end{eqnarray}
where ${\bf{U}}_{\rm{train}}$ is the training dataset and ${\bf{F}}$ is the function that encodes the transformation of the input into the outcome of measuring the states of the reservoir neurons. The weight matrix $\bf{W}$ is then optimized through a training process to align the neural network output ${\tilde{\bf{Y}}}$ with the target vector ${{\bf{Y}}}_{\rm{train}}$. The weight matrix $\bf{W}$ can be calculated as
\begin{eqnarray}
{\bf{W}}={\tilde{{\bf{Y}}}}_{\rm{train}}{\bf{F}}^{\dagger} ({\bf{U}}_{\rm{train}})\,,
\label{eq:RC8}
\end{eqnarray}
where ${\bf{F}}^{\dagger}$ is the Moore–Penrose pseudo inverse of the function ${\bf{F}}$. Then, at the stage of exploitation of the reservoir, the optimised weight matrix is applied to a test dataset ${\bf{U}}_{\rm{test}}$ as
\begin{eqnarray}
{\bf{Y}}_{\rm{test}}={\bf{WF}}({\bf{U}}_{\rm{test}})\,.
\end{eqnarray}

In an idealised scenario, the prediction made by a trained reservoir should coincide with the target dataset (also called the ground truth). However, in practice the output of the RC system deviates from the target. In the literature on reservoir computing, the forecast-target deviation is often quantified using the root-mean-square error (RMSE) that can be calculated as
\begin{eqnarray}
{\rm{RMSE}}=\frac{1}{{y}_{\max }-{y}_{\min }}\sqrt{\frac{\mathop{\sum }\nolimits_{i}^{N}{({y}_{i}-{\tilde{y}}_{i})}^{2}}{N}}\,,
\end{eqnarray}
where $N$ is the total number of data points taken into account in the calculation of RMSE, $y_i$ is the actual target value for the $i$th data point and $\tilde{y}_i$ is the value predicted by the RC system. The range of the target values is accounted for by the values of $y_{\max}$ and $y_{\min}$. In addition, we evaluate the performance of the reservoir using the figure-of-merit called the accuracy, which is calculated as the ratio of the correct predictions to the total number of predictions made by the RC system:
\begin{eqnarray}
Accuracy = \frac{\mathop{\sum }\nolimits_{i}^{N_{eqv}}{|{y}_{i}-{\tilde{y}}_{i}|<\epsilon}}{N_{eqv}}\times 100\,, 
\end{eqnarray}
The agreement percentage is calculated such that the difference between the correct predictions and the target is less than $\epsilon=10^{-2}$.

\section{Results and Discussion\label{sec:Test}}
\subsection{Task classification}
To evaluate the accuracy of the predictions made by our RC system, we task it with a series of test problems. The first problem, which was used to assess the performance of several previously developed quantum RC systems \cite{Rio17, Dud23}, involves a binary categorisation a synthetic waveform composed of randomly generated sinusoidal and square pulses. In this task, the output of the reservoir is expected to be 0 (1) when an input point corresponds to the square (sinusoidal) portion of the test waveform. This task is designed to test the memory capacity and nonlinearity of the reservoir \cite{Dud23}.

\begin{figure}[H]
\centering
\includegraphics[angle=0,width=0.8\columnwidth]{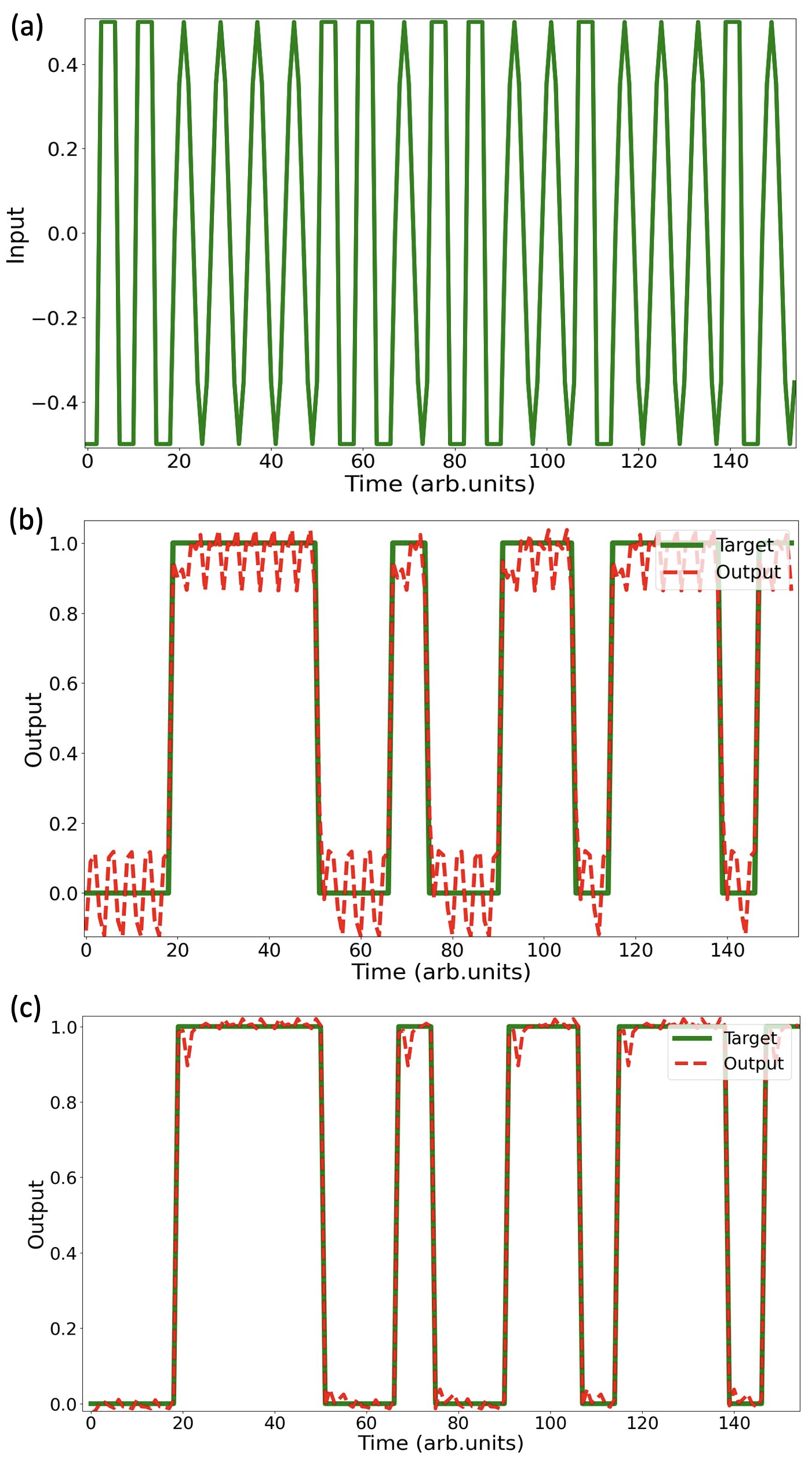}
\caption{{\bf(a)}~Input data generated from a random array representing either a sinusoidal or square waveform. Results of the classification of the sinusoidal and square waveform by the reservoir with {\bf(b)}~8 and {\bf(c)}~16 neurons. The target is shown in solid green line and the reservoir prediction in dashed red line.}
\label{f3}
\end{figure}

The input used in this test corresponds to a time series obtained by sampling the discrete data points from an array representing either a sinusoidal or square waveform, the so-obtained dataset is split into two parts: one part is used for training and another one for testing (Figure~\ref{f3}a). 

\begin{figure}[th]
\centering
\includegraphics[width=0.8\columnwidth]{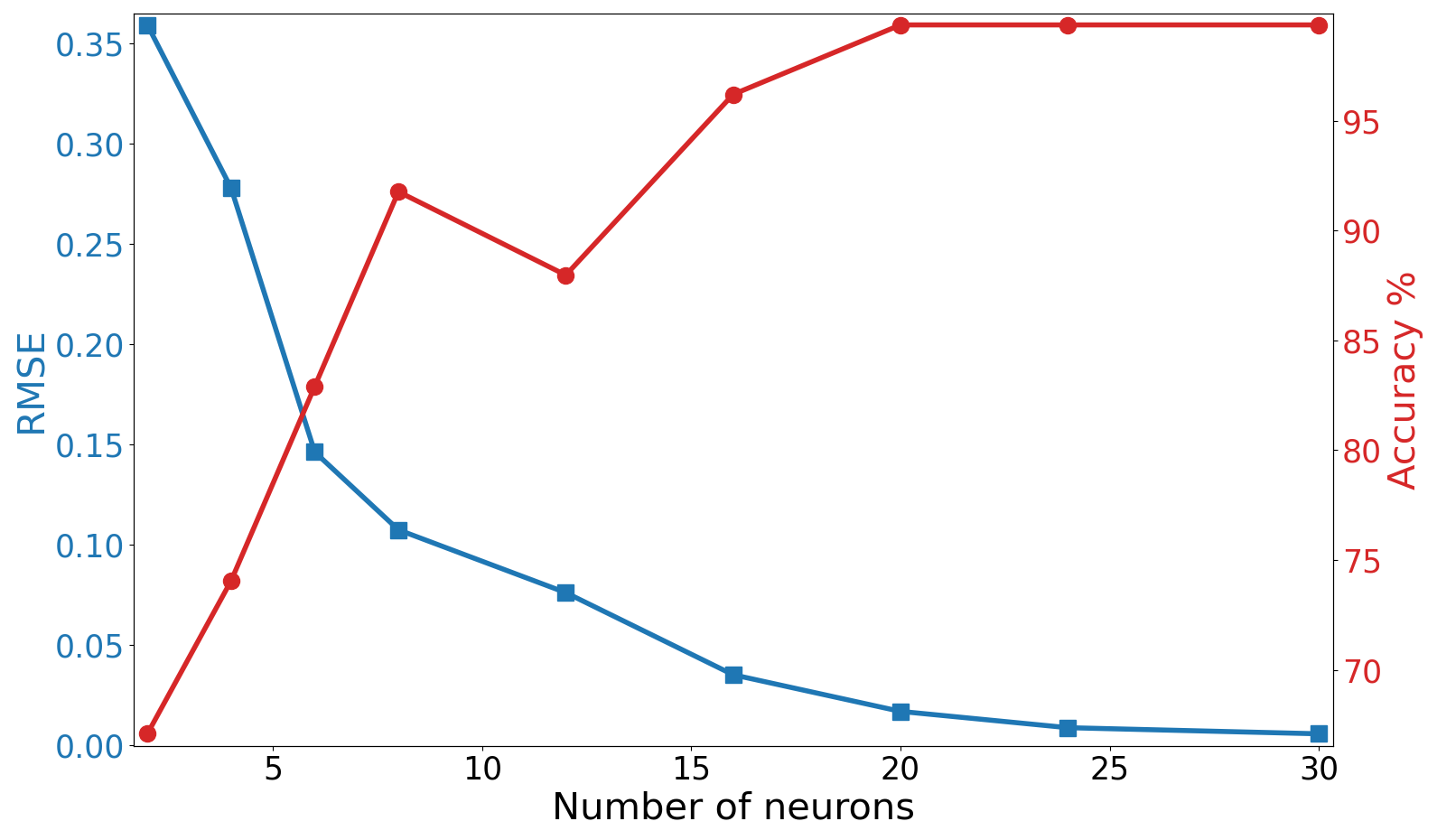}
\caption{The root-mean-square error (RMSE) (the blue square markers) and the accuracy (the red circular markers) obtained for the sinusoidal-square waveform classification task as a function of the number of neurons in the reservoir.}
\label{f4}
\end{figure}

The performance of the RC system configured to have 8 and 16\,neurons is evaluated in Figure~\ref{f3}b and Figure~\ref{f3}c, respectively, where the solid green line denoted the target but the dashed red line corresponds to the output of the reservoir. We note that the target signal is presented purely for the comparison of the reservoir output with the ground truth, i.e.~the reservoir is not presented with the target data at the exploitation stage. We can see that the ability of the reservoir to classify the input improves as the number of neurons is increased. The reservoir with 16\,neurons can complete the test task with the accuracy of $99.7\%$. 
To gain a deeper insight into the dependence of the reservoir performance on the number of neurons, Figure~\ref{f4} plots the RMSE and accuracy as a function of the number of neurons employed in the classification task. Notably, the reservoir with just 8\,neurons demonstrates an accuracy of approximately $90\%$ and RMSE of approximately 0.1. However, increasing the number of neurons to 16 results in a significant improvement, yielding a remarkable $99.7\%$ accuracy and a reduced error of $4 \times 10^{-3}$. Importantly, any further increase in the number of neurons beyond 16 results in just a marginal increase in the accuracy, indicating that 16\,neurons is an optimal configuration that enables the reservoir to make feasible forecasts using less energy and resources.

The results presented in Figure~\ref{f4} demonstrate the advantage of the proposed quantum reservoir over a classical one: the quantum reservoir enables us to achieve significant accuracy using a small number of neurons. Indeed, a traditional RC algorithm typically requires more than 1000\,neurons to produce a plausible forecast (see, e.g., Refs.~\cite{Luk09, Luk12} and the computational code that accompanies them). In turn, physical counterparts of the traditional RC algorithms require at least 40\,neurons to complete test tasks of comparable complexity \cite{Wat20, Mak23_EPL}. We also used the quantum RC software that accompanies the discussion in Ref.~\cite{Dud23} to undertake the same task as in Figure~\ref{f3}. To enable the correct benchmarking, that computational code was run on the same workstation computer equipped with the same version of Python language used to implement our reservoir model. We revealed that, for the same length of the training dataset, our reservoir accomplished the task in 9\,seconds compared with 50\,seconds required by the model reported in Ref.~\cite{Dud23}. 

\begin{figure}[th]
\centering
\includegraphics[width=0.8\columnwidth]{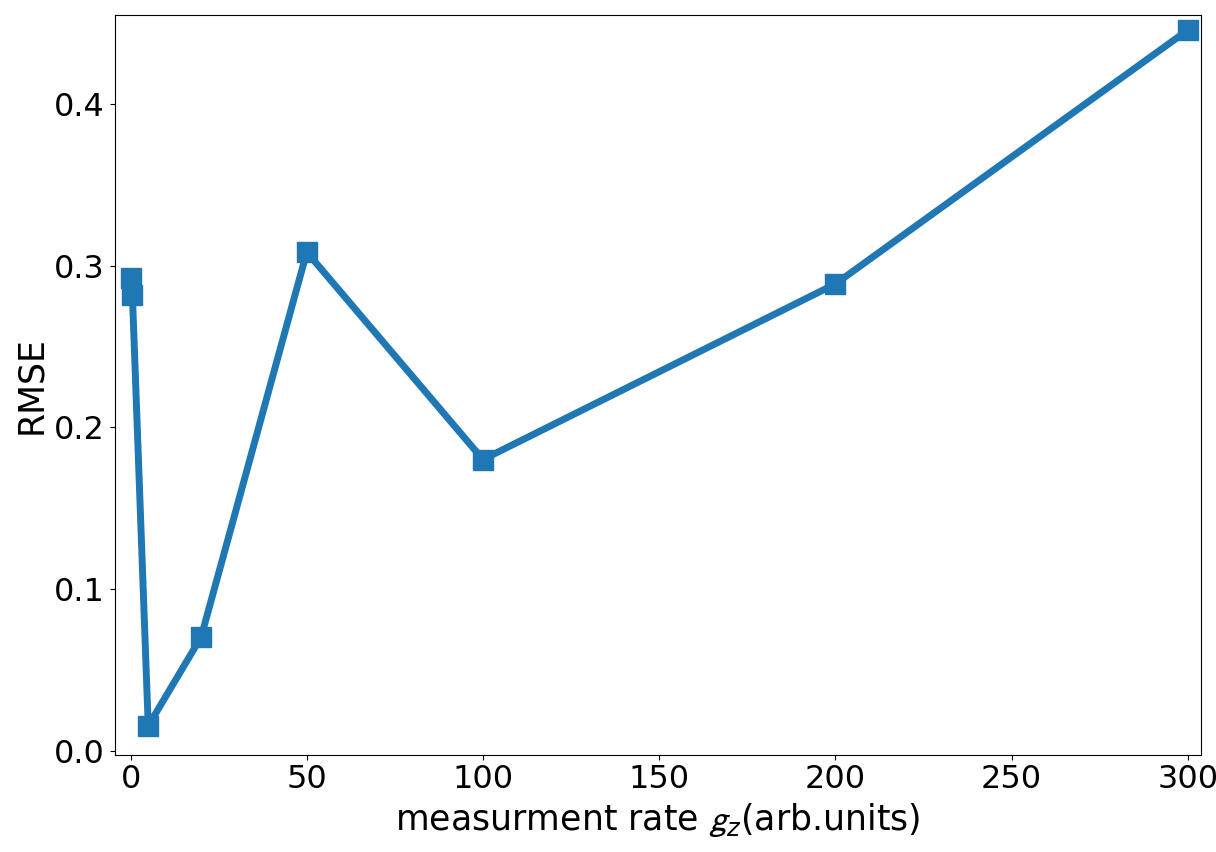}
\caption{Performance of the reservoir as the function of the measurement rate. The optimal RMSE is achieved at the measurement rate of 5\,arb.~units.}
\label{f5}
\end{figure}

Figure~\ref{f5} reveals that a superior performance of our reservoir compared with the previously proposed quantum reservoirs stems from the use of the feature of measurement-controlled dynamics: the value of RMSE is a function of the measurement rate, with an optimal performance at $g_z = 5$\,arb.~units. To obtain a bigger picture, we computed RMSE for the larger values of the measurement rate, demonstrating that a high rate does not result in better performance (yet, we note that low rate (continues measurement) computations are time consuming). Further analysis of the physical properties of the reservoir and their impact on its computational performance will be reported elsewhere.

\subsection{Chaotic times-series forecasting}
The second test task used to evaluate the performance of our RC system consists in predicting a Mackay-Glass time series (MGTS) that is generated solving the delay differential equation \cite{Mac77}
\begin{eqnarray}
  \dot{x}_{_{MGTS}}(t)
  &=&\beta_{_{MGTS}}\frac{x_{_{MGTS}}(t-\tau_{_{MGTS}})}
      {1+x_{_{MGTS}}^{q}(t-\tau_{_{MGTS}})}-\gamma_{_{MGTS}}x_{_{MGTS}}(t)\,,
  \label{eq:MGTS}
\end{eqnarray}
where overdot denotes differentiation with respect to time and $\tau_{_{MGTS}}=17$, $q=10$, $\beta_{_{MGTS}}=0.2$ and $\gamma_{_{MGTS}}=0.1$ \cite{Luk12}. The generated time series is then split into two parts used for at training and testing stages, respectively.

In this test task, the reservoir operates in the generative mode, also known as the free-running forecast, where the output produced by a trained reservoir in the previous time step serves as an input at the next time step, i.e.~$u_{n+1} = y_n$ \cite{Luk12}. Hence, the reservoir acts as a self-generator during this phase \cite{Sho23}. We stress that we deliberately choose to test the reservoir in the generative mode because, as shown in Refs.~\cite{Luk09, Luk12, Mak21_ESN, Mak23_EPL, Mak23_review, Mak24_reservoir}, demonstrations of the operation in the predictive mode are technically straightforward. While the operation in the generative mode is a more challenging task, the practical importance of generative reservoirs is typically much higher since they can be used to solve a wide range of problems concerned with the prediction of difficult to analyse processes such as behaviour of financial markets and variations of climate \cite{Mak23_review}.
\begin{figure}[H]
\centering
   \includegraphics[width=0.87\columnwidth]{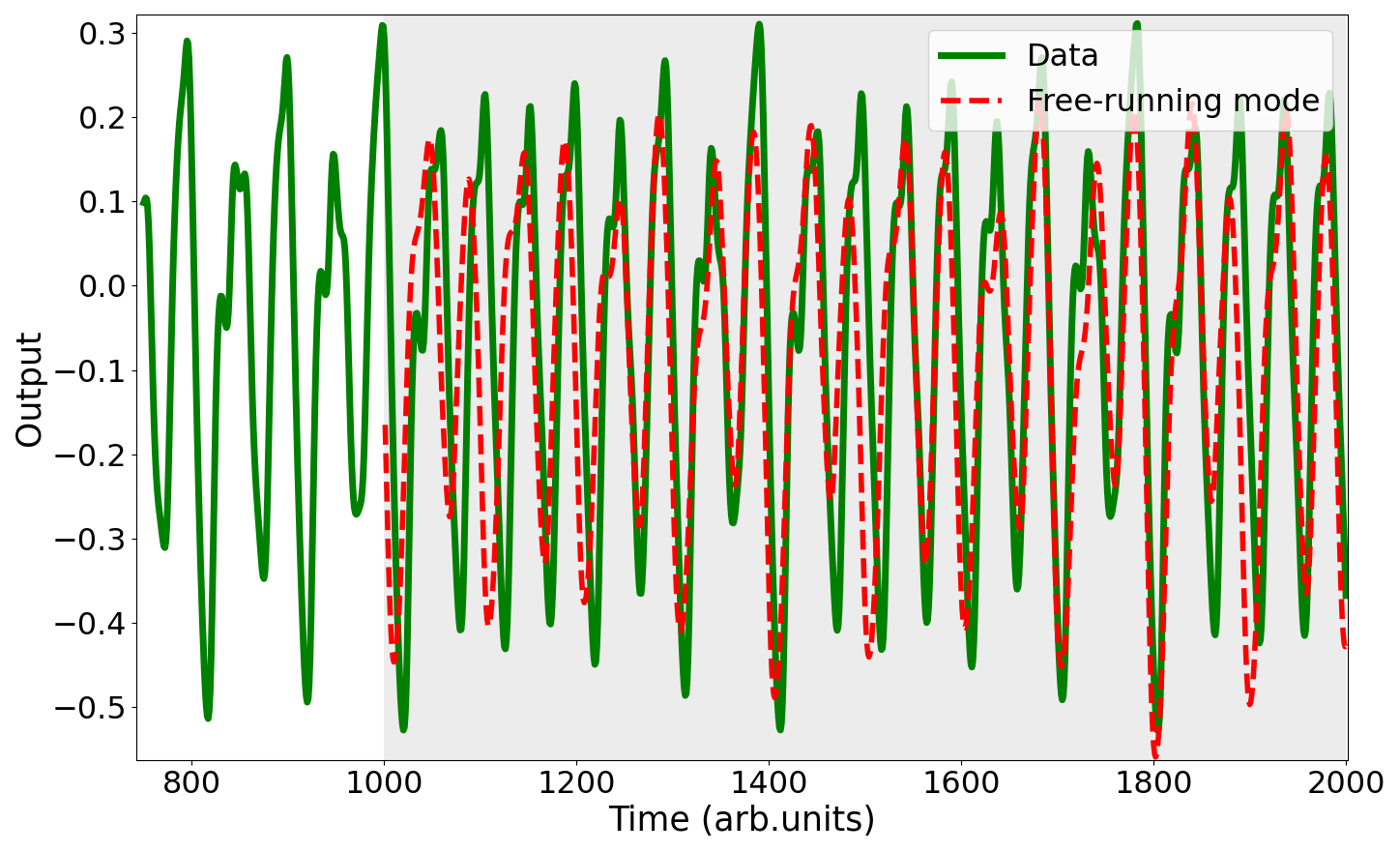}
\caption{Generative mode operation exemplified by a free-running forecast of MGTS. In this figure, we compare the output of the reservoir (the dashed-red line) with the target MGTS (the solid green line). The reservoir has 16 neurons and it was trained on several cycles of MGTS variations. Note that the reservoir was not presented with the ground truth MGTS data to make the forecast. The comparison with the ground truth is needed only to evaluate the accuracy of the forecast.}
\label{f6}
\end{figure}

Figure~\ref{f6} demonstrates the result of the free-running forecast made by the reservoir with 16\,neurons. The forecast future evolution of MGTS is denoted by the red dashed line and it starts at the instant of time of 1000\,arb.~units. The green solid line corresponds to the ground truth. We can see that the RC system correctly reproduces the general pattern of MGTS, though it misses some minor features of the ground truth time series.

A free-running forecast of MGTS is as one of the standard benchmarking problems used to evaluate the performance of the traditional RC algorithm \cite{Luk12}. Using the computer code from Ref.~\cite{Luk12} we established that a traditional RC system can produce a free-running forecast of comparable accuracy only when the reservoir contains more than 500\,neurons (the other hyperparameters of the traditional RC system used in this comparison were the leaking rate $\alpha=0.3$ and the ridge parameter $\lambda=10^{-6}$; the spectral radius was computed according to the procedure outlined in Refs.~\cite{Luk09, Luk12}). Thus, we conclude that a reservoir operating using the principles of quantum mechanics can requires a much smaller number of neurons compared with a classical reservoir. 
\begin{figure}[th]
\centering
\includegraphics[width=0.8\columnwidth]{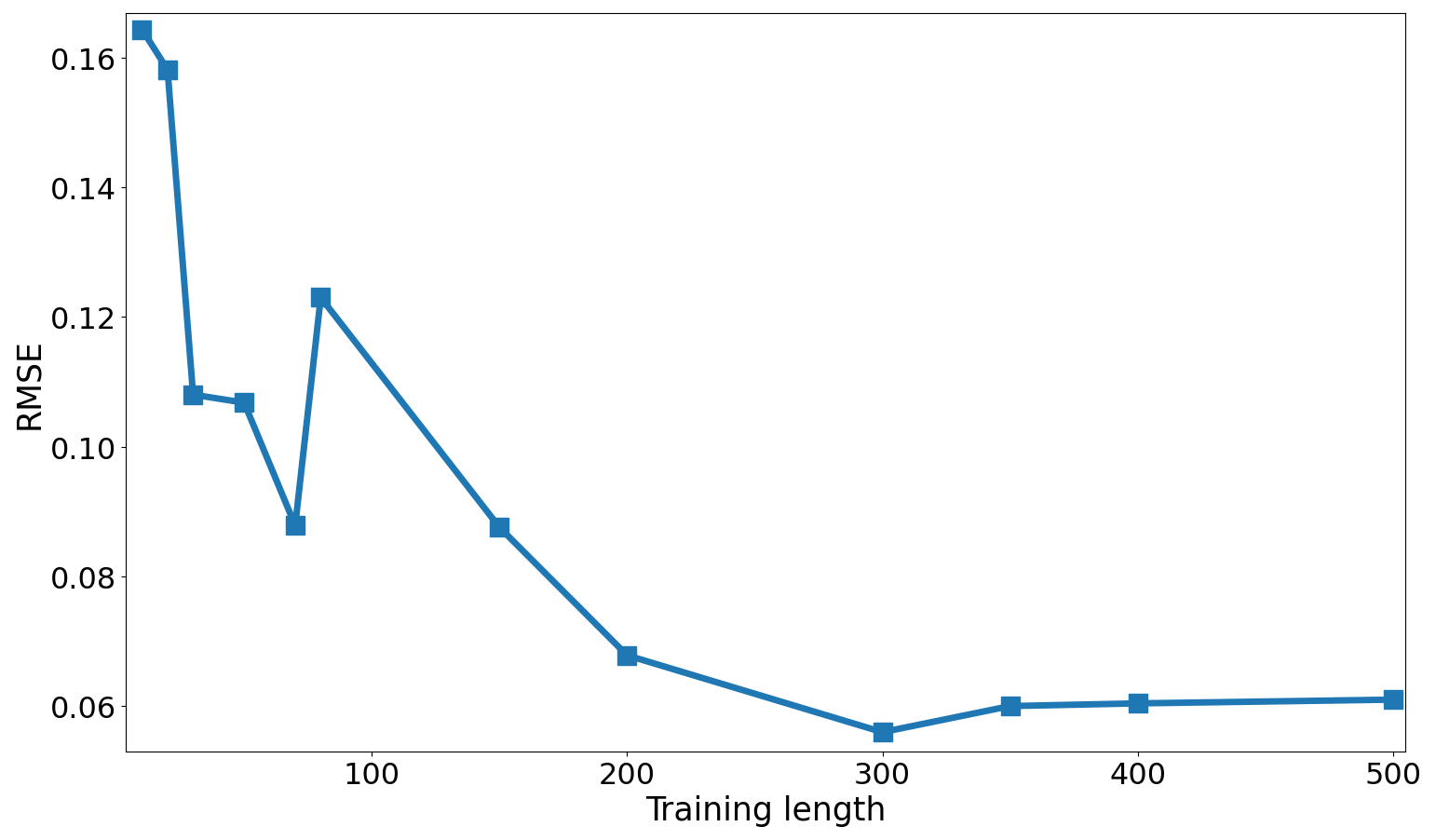}
\caption{RMSE plotted as a function of the length of the training MGTS dataset.}
\label{f7}
\end{figure}

In a previous work \cite{Mak24_reservoir}, we also demonstrated that the correct operation of a traditional RC algorithm is possible mostly using relatively long training data sets. In fact, the traditional RC system implemented following the procedure outlined in Refs.~\cite{Luk09, Luk12} requires approximately 1000 training MGTS data points. In Figure~\ref{f7} we plot RMSE as a function of the length of the MGTS data set used to train our RC system. We can see that a reasonable accuracy can be reached with a minimum of 300 training data points. Any further increase in the length of the training data set does not result in significant increase in the accuracy. This results mirrors our previous observation made in Ref.~\cite{Mak24_reservoir}: an RC system based on the physical principles can be trained using shorter data sets compared with the dataset length required to train a traditional RC system. 

\subsection{Damped harmonic oscillator prediction}
In this section, we demonstrate the ability of the proposed RC system to learn and predict real-life physical phenomena. As a test task, the RC system is presented with several periods of oscillation of a harmonic oscillator, a system that, when displaced from its equilibrium position, experiences a restoring force that is proportional to the displacement. In the presence of damping, depending on the friction coefficient, the system oscillates with a frequency that is lower than that in the undamped case but the amplitude of oscillations decreases with time.

Despite its relative simplicity, this test task is non-trivial since it requires any RC system to adjust to constantly changing input conditions, also requiring strong output feedback needed to generate oscillations (for this reason, a conceptually similar test problem called frequency generator was used used to demonstrate the operation of the traditional RC algorithm by its creators \cite{Scholarpedia, Mak23_review}). Moreover, undertaking this specific task presents a considerable challenge to the reservoir due to a two-parameter space nature of the problem: the RC system should be trained to accurately simulate the dynamics of both frequency and amplitude.

\begin{figure}[H]
\centering
\includegraphics[width=0.85\columnwidth]{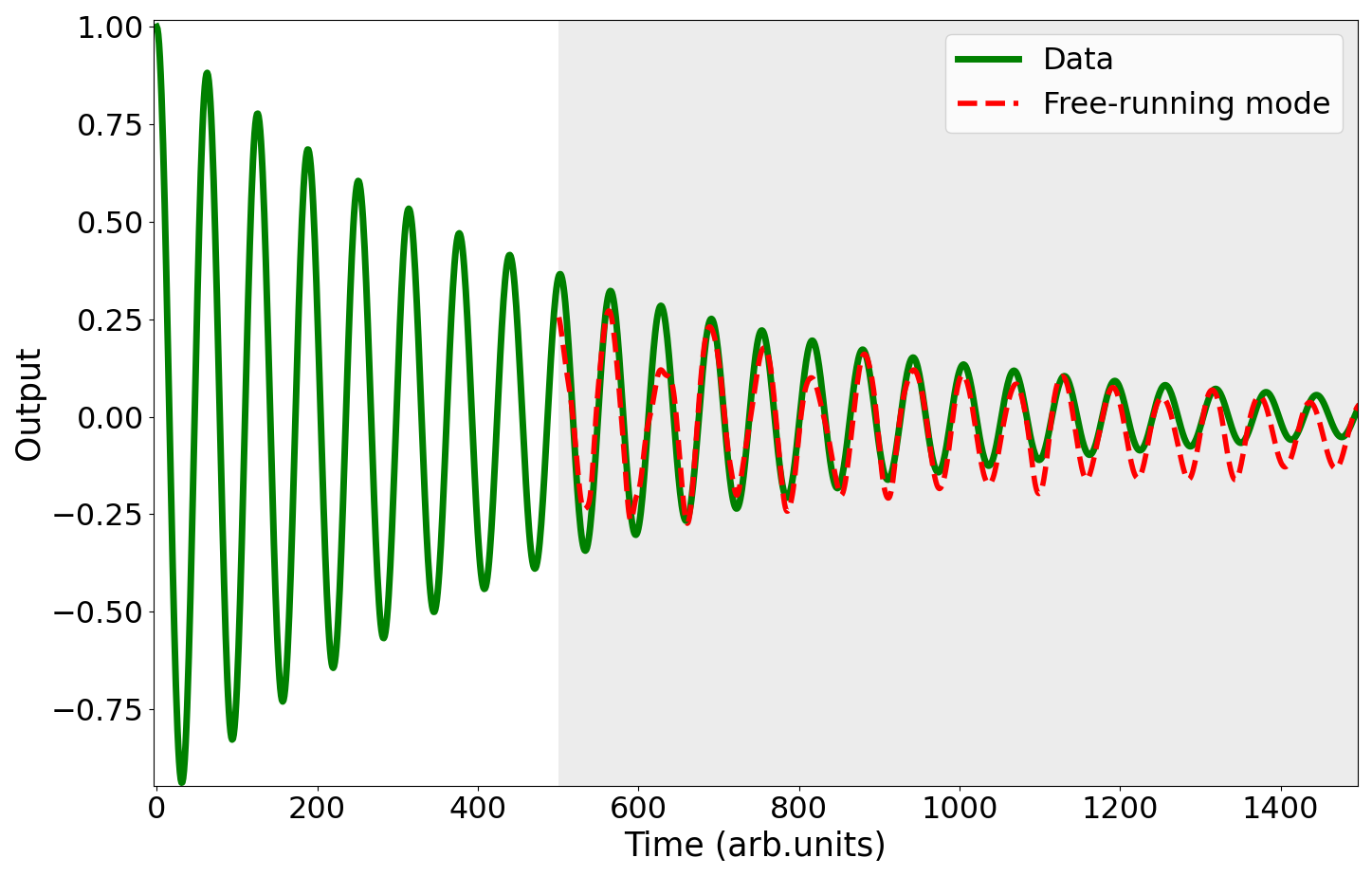}
\caption{Output of the RC system trained to predict a damped harmonic oscillator. The free-running forecast made by the the RC system is denoted  by the dashed red line. The solid green line denotes the ground truth.}
\label{f8}
\end{figure}

In Figure~\ref{f8} we plot the forecasting made by the reservoir trained on a damped oscillator dataset. We observe a good agreement with the ground truth over the first few periods. Then we can see certain deviations of the free-running forecast from the natural dynamics of the oscillator. Despite this artefact, in the following section we demonstrate that the ability of the RC system to predict a harmonic oscillator can find practical applications in the framework of the paradigm of approximate computing.

\section{Discussion}
We have demonstrated that the proposed quantum RC system can successfully undertake several challenging test tasks using a very small number of neurons compared with the traditional RC systems. Of course, software that implements the traditional RC algorithm can make a more accurate forecast using several thousands of neurons. However, such a computation will require a high-performance workstation computer that will need to be run for several hours or even days to find the optimal set of hyperparameters. We also established that our quantum RC system outperforms the other quantum-physical reservoirs in terms of computational resources needed to solve the standard test problems. 

These characteristics make our RC system suitable for applications in approximate computing. Yet, our findings can also be used in embedded systems that typically have limited processing power and memory compared to general-purpose computers, often requiring low power consumption.

For example, while, admittedly, the forecast made by the RC system in Figure~\ref{f8} is not ideal when compared with the ground truth, the demonstrated operation of the reservoir holds the potential to significantly facilitate certain computational and experimental procedures, thereby fitting into the paradigm of approximate computing. Indeed, a quantum RC system trained using a damped harmonic oscillator may become a valuable tool for the investigation of individual quantum emitters that serve as a fundamental building block of many emerging devices \cite{Moc12, Bas20}. From the computational point of view, the design of such devices requires the calculation of quality factors and decay rates of photonic resonators integrated with waveguiding structures, which is typically accomplished using complex numerical techniques, including the finite-difference time-domain (FDTD) method \cite{Taflove}, or sophisticated analytical models \cite{Yan15}.  

The FDTD method is popular in the field of photonics since it enables one to accurately represent the geometry of optical resonators and waveguides while taking into account the optical properties of materials such as dispersion, nonlinearity and absorption in wide range of optical frequencies of interest. However, due to the time domain nature of its algorithm, calculations of quality factors and decay rates of realistic photonic structures require one to run FDTD software for a long time to reach a steady state regime.

While several approaches intended to decrease the computational effort have been proposed \cite{Guo01}, a typical FDTD simulation requires at least several hours of CPU time of a high-performance workstation computer. Our RC system can be used to further decrease the calculation time: it can be trained on data obtained after a relatively short FDTD simulation (typically it requires approximately 30\,minutes to generate such data) and then used to predict the further temporal evolution. Since such a hybrid calculation will take about one hour in total, the user can save from two to eight hours of CPU time per simulation. Given that a typical computational research project involves several tens of independent simulation runs, the application of the RC system can potentially save up to 800 CPU hours. Of course, using this combined approach one needs to take into account the imperfection of forecasts made by the RC system. On the other hand, even well-designed FDTD simulations are not free of numerical artefacts that can be of the same order of magnitude as the imperfection of the forecast. Yet, in some simulations the FDTD method can suffer from late time instabilities that may not allow the user to run the software until the steady state regime is reached \cite{Dou02}. This problem can also be resolved using the RC system.   

Researchers conducing experimental work encounter similar problems during the analysis of raw data. For example, this is the case of research on nanodiamonds containing fluorescent nitrogen-vacancy (NV) centres that are essential for biomedical imaging and sensing \cite{Rei19}. The rates of radiative and non-radiative decay of the excited NV centre states are affected by surface proximity effects \cite{Rei17}. The analysis of these decay rates shapes the research in the field, relying on different models that produce best fits of experimental data \cite{Rei17, Rei19}. In many cases, the fitting procedure needs to be done manually. The RC system can be used to optimise the data fitting procedure, helping the researchers predict the decay rates using noisy or incomplete experimental data sets. A similar RC-based fitting procedure can also be used to investigate the oscillation of many other physical systems, including magnetic gas sensors \cite{Mak_gas} and nonlinearly oscillating gas bubbles in liquids \cite{Mak22_Bio}.     

In many practical applications, including the aforementioned areas of biomedical imaging and sensing, data processing needs to be done using portable and miniaturised systems. Usually, such systems have limited computational resources and designed to consume low power. The quantum reservoir proposed in this paper can be used to optimise the performance of such systems. The large computational power is required to perform operation with matrices whose size depends on the number of neurons in the reservoir. Since our quantum RC system uses a small number neurons, the matrices associated with its algorithm are also small. In our recent work \cite{Mak24_reservoir}, we demonstrated that software that implements an RC system that requires small matrices can be run on inexpensive microcontrollers. We also showed that microcontroller-based RC systems can be integrated with various sensors and actuators, thereby enabling one to implement an approximate computing scheme in a field experiment and conduct calculations on board of an unmanned aerial vehicle (UAV) \cite{Por20}. An on-board RC system can also help an UAV to recognise other drones \cite{Hen22_1}.   

\section{Conclusions}
We have proposed and theoretically validated a computational reservoir system that operates using the dynamics of a probed atom in a cavity and relies on the control of the quantum measurement rate. Benchmarking the performance of the reservoir using several challenging test problems, we demonstrated that feasible forecasts can be made using just 16\,artificial neurons compared with approximately 1000\,classical artificial neurons require for the operation of a traditional reservoir computing system. We also showed that our quantum reservoir produces accurate results even when it is trained on relatively short training datasets.

While the performance of the traditional reservoir can be improved, until a certain point, by increasing the number of neurons and optimising the set of classical reservoir hyperparameters, including the leaking rate and spectral radius, this procedure will require using an expensive and difficult to access high-performance computer. Subsequently, given that the quantum reservoir can produce usable results with a small number of neurons, it is plausible that it may be used to solve many practical problems in the framework of the paradigm of approximate computing.     

Yet, compared with the other quantum and classical reservoir systems, software that implements our reservoir can be run on ordinary desktop and laptop computers equipped with modest computational resources. This quality makes its possible to utilise our reservoir in embedded and portable systems, including various type of sensors and actuators that can be integrated with autonomous vehicles and robots.    

\vspace{6pt} 




\authorcontributions{Conceptualization, writing, editing, and discussion A.~H.~Abbas and Ivan S.~Maksymov; methodology; software; and validation, A.~H.~Abbas; supervision, Ivan S.~Maksymov. All authors have read and agreed to the published version of the manuscript.}

\funding{This research received no external funding.}

\institutionalreview{Not applicable.}

\informedconsent{Not applicable.}

\dataavailability{The code used in this study is available from the corresponding author upon reasonable request.} 


\conflictsofinterest{The authors declare no conflicts of interest.} 



\abbreviations{Abbreviations}{
The following abbreviations are used in this manuscript:\\

\noindent 
\begin{tabular}{@{}ll}
CPU & central processing unit\\
FDTD & finite-difference time-domain\\
MGTS & Mackay-Glass time series\\
NV & nitrogen-vacancy \\
RC & reservoir computing\\
RMSE & root-mean-square error\\
UAV & unmanned aerial vehicle\\
\end{tabular}
}

\begin{adjustwidth}{-\extralength}{0cm}

\reftitle{References}


\externalbibliography{yes}
\bibliography{refs}

%


\end{adjustwidth}
\end{document}